\documentclass[journal,twoside,web]{ieeecolor}
\usepackage{tmi}
\usepackage{cite}
\usepackage{amsmath,amssymb,amsfonts}
\usepackage{algorithmic}
\usepackage{graphicx}
\usepackage{textcomp}
\usepackage{booktabs}   
\usepackage{subcaption}
\usepackage{makecell}
\usepackage[dvipsnames]{xcolor}
\usepackage{arydshln}

\usepackage{algorithm}

\graphicspath{ {images/} }

\newif\ifdraft
\drafttrue

\ifdraft
\def\HB#1{{#1}}
\def\HBc#1{}  
\def\HBd#1{\textcolor{purple}{\textit{[deleted: #1]}}}  

\def\D#1{{#1}}
\def\Dc#1{}  
\def\Dd#1{\textcolor{purple}{\textit{[deleted: #1]}}}  

\def\Pc#1{\textcolor{blue}{\textit{\textsf{ \small [PS: #1]}}}}  
\def\Pd#1{\textcolor{purple}{\textit{[deleted: #1]}}}  

\def\ACc#1{\textcolor{brown}{\textit{\textsf{ \small [AC: #1]}}}}  
\def\ACd#1{\textcolor{gray}{\textit{[AC deleted: #1]}}}  

\else

\def\HB#1{#1}
\def\Hc#1{}  
\def\HBc#1{}  
\def\Hd#1{}  
\def\HBd#1{}  

\def\D#1{#1}
\def\Dc#1{}  
\def\Dd#1{}  

\def\Pc#1{}  
\def\Pd#1{}  

\def\ACc#1{}  
\def\ACd#1{}  

\fi

\def\BibTeX{{\rm B\kern-.05em{\sc i\kern-.025em b}\kern-.08em
    T\kern-.1667em\lower.7ex\hbox{E}\kern-.125emX}}
\begin{document}
\title{Learning Spatio-Temporal Model of Disease Progression with NeuralODEs \HB{from Longitudinal Volumetric Data}}
\author{Dmitrii Lachinov, Arunava Chakravarty, Christoph Grechenig, Ursula Schmidt-Erfurth, Hrvoje Bogunovi\'c

\thanks{Manuscript submitted Oct, 14, 2022. This work was supported in part by the Christian Doppler Research Association, Austrian Federal Ministry for Digital and Economic Affairs, the National Foundation for Research, Technology and Development, and Heidelberg Engineering.}
\thanks{Dmitrii Lachinov and Hrvoje Bogunovi\'{c} are with the Christian Doppler Lab for Artificial Intelligence in Retina, Department of Ophthalmology and Optometry, Medical University of Vienna, Austria (e-mail: dmitrii.lachinov@meduniwien.ac.at, hrvoje.bogunovic@meduniwien.ac.at).}
\thanks{Arunava Chakravarthy, Christoph Grechenig, and Ursula Schmidt-Erfurth are with the Department of Ophthalmology and Optometry, Medical University of Vienna, Austria.}
}

\maketitle

\begin{abstract}
Robust forecasting of the future anatomical changes inflicted by an ongoing disease is an extremely challenging task that is out of grasp even for experienced healthcare professionals. Such a capability, however, is of great importance since it can improve patient management by providing information on the speed of disease progression already at the admission stage, or it can enrich the clinical trials with fast progressors and avoid the need for control arms by the means of digital twins.
In this work, we develop a deep learning method that models the evolution of age-related disease by processing a single medical scan and providing a segmentation of the target anatomy at a requested future point in time. Our method represents a time-invariant physical process and solves a large-scale problem of modeling temporal pixel-level changes utilizing NeuralODEs. In addition, we demonstrate the approaches to incorporate the prior domain-specific constraints into our method and define temporal Dice loss for learning temporal objectives.
To evaluate the applicability of our approach across different age-related diseases and imaging modalities, we developed and tested the proposed method on the datasets with 967 retinal OCT volumes of 100 patients with Geographic Atrophy, and 2823 brain MRI volumes of 633 patients with Alzheimer's Disease.
For Geographic Atrophy, the proposed method outperformed the related baseline models in the atrophy growth prediction. For Alzheimer's Disease, the proposed method demonstrated remarkable performance predicting the brain ventricle changes induced by the disease, achieving the state-of-the-art result on TADPOLE challenge.

%
\end{abstract}


\section{Introduction}
\label{sec:introduction}

%


The progress in the medical imaging techniques brought vast possibilities for the diagnosis and monitoring of patients with age-related diseases. Tools such as Optical Coherence Tomography (OCT) or Magnetic Resonance Imaging (MRI) are the current clinical gold standard and contribute to an overwhelming number of diagnostic findings. The amount of information within these scans, however, is so large, that it requires many years of training and practice to interpret them. Yet, we believe, a lot stays hidden from expert eyes. For instance, we hypothesize that the track of the natural, age-related disease progression is determined by the subtle physiological changes which are captured and encoded by medical scans. The class of computational methods trying to extract such information is called Disease Progression Models (DPMs), which can, e.g. calculate the expected future change in the target quantitative biomarkers. A more sophisticated \HB{class of} methods can forecast morphological change with respect to the shape and size of an anatomical structure or lesion that is expected to occur in the future, from an image acquired at the time of screening. The task of modeling future changes is extremely challenging as the early biomarkers that indicate either a fast or slow speed of progression are often very subtle and subclinical, i.e., not well-understood in clinical practice.

DPMs have already been employed as decision support and as tools complementing phase II and III clinical trials~\cite{dpm_role,phase2trial,phase2trial2}. They typically model the evolution of \textit{a set of} predefined biomarkers over the course of the disease and map it to the global \D{timeline} \D{of disease evolution}, where the biomarkers are typically extracted from clinical reports and imaging data. However, such biomarkers relevant to model the disease progression have to be predetermined by the human experts, which is often \D{a non-trivial task}, \HB{and likely does not cover all the revelant ones}. Thus, the search for predictive biomarkers is currently a complicated and tedious procedure.



The state-of-the-art methods that automatically extract features from the input volume also have some weaknesses, as some are designed for forecasting of fixed intervals only; others are time-variant systems, as the same input volume can produce different outcomes at different times. 
In this work, we propose a novel disease progression model capable of learning \D{directly} from longitudinal imaging data to predict the target area or volume change from a single scan, acquired at baseline visit. The proposed model not only predicts \D{continuous-in-time} changes in the target size, but it is time-invariant and inherently interpretable, as it simultaneously produces a detailed segmentation map of the current and the future target anatomy.

\subsection{Clinical background}
We focus on developing a DPM method for two clinically-relevant scenarios: (1) modeling the growth of Geographic Atrophy (GA) from longitudinal 3D OCT volumes of the retina to monitor Age-Related Macular Degeneration (AMD), 
and (2) modeling Alzheimer's Disease (AD) progression by capturing the morphological changes in the brain ventricles from longitudinal 3D MRI volumes of the brain.

\subsubsection{Geographic Atrophy}
\label{sec:ga}
AMD is a leading cause of blindness, affecting 8.7\% of the world population~\cite{WONG2014e106}. In the 80+ age group, this number grows to 25\%, with the prevalence of advanced AMD of 3.3\%. AMD in its late stages can either be \textit{dry} characterized by the presence of Geographic Atrophy (GA) or \textit{wet} characterized by the presence of \HB{Macular} Neovascularization.
OCT has become a gold standard modality for imaging the retina in the clinical setting due to its non-invasive 3D acquisition and widespread availability in clinics. It acquires a detailed volumetric scan of the retina made of a series of cross-sectional slices or BScans (Fig. \ref{fig:data_example}). On a BScan, GA is represented as an atrophy of the outer retinal tissues~\cite{GUYMER2020}. Thus, each column (AScan) in a BScan can be denoted as atrophic or not. This makes the GA lesion to be 2D in nature (Fig. \ref{fig:data_example}), with its size typically measured as an en-face area in mm$^2$.
In contrast to the late-wet stage of AMD, the late-dry stage \HB{is more prevalent, affecting around 5 Mio worldwide}, and has no approved treatments yet. However, very recently multiple drugs have passed phase 3 trials, most prominently pagcetacoplan~\cite{filly_study} and avacincaptad pegol~\cite{Jaffe2021}.
Thus, modeling the progression of GA plays a critical role in clinical research to understand the pathophysiology, \HB{and the simulation of the future atrophy progression is expected to be fundamental to patient management in order to identify fast progressors that will need the treatment the most}.

\subsubsection{Alzheimer's Disease}
\label{sec:alzheimer}
\begin{figure*}[tbh]
\centering
\includegraphics[width=0.9\linewidth]{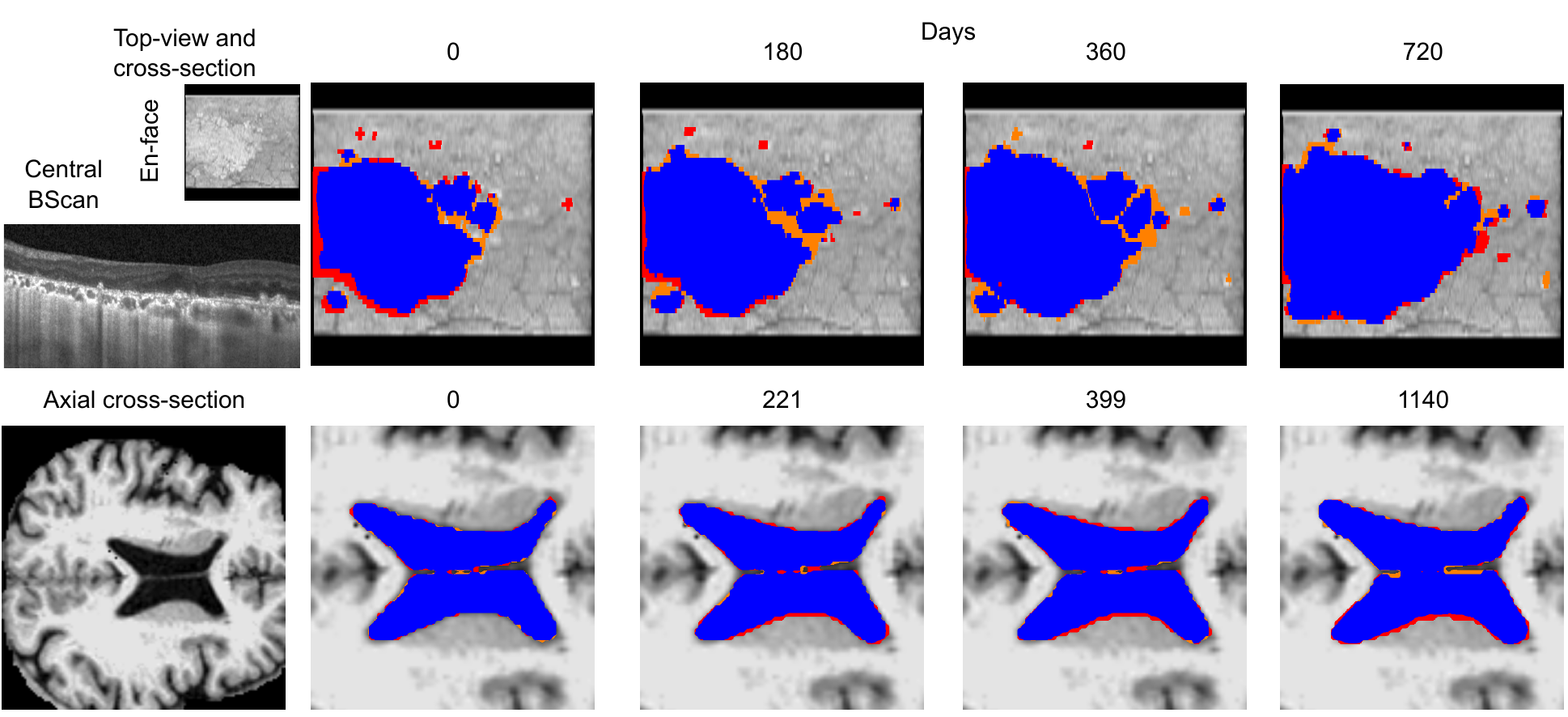}
\caption{The proposed method predicts the future target anatomy segmentation from a single volumetric scan acquired at baseline (day 0). The top row represents the en-face projections of the volumetric 3D OCT scan with the atrophy segmentation predicted at different future visits, denoted by the number of days since the baseline visit. The raw baseline OCT volume is shown as an en-face projection and with a central BScan (cross-section). Note that new islands of atrophy form in the central right and top left locations of the scan.
The bottom row represents an axial slice of the brain T1 MRI cropped around the ventricles, with their predicted segmentations at future time points. The growth in a 2D slice is marginal, but it is substantial in 3D. In both rows, true positives (blue), false positives (orange), and false negatives (red) are depicted. 
}
\label{fig:data_example}
\end{figure*}

AD is a gradually progressing, irreversible neurodegenerative disorder which is the primary of Dementia.
Around 50 million patients suffer from this impairment worldwide, with the projected number of patients effectively doubling each 5 years and reaching 152 million by 2050 \cite{ad_review}. Currently, only symptomatic treatment is available.
Volumetric MRI of the brain is commonly employed in the clinical examination of AD to detect the early pathological changes in the cortex, hippocampus and ventricles. Recent studies by~\cite{Ott2010,Nestor2008} have correlated the ventricle volume extracted from MRI images to the progression of AD, thereby making it a potential biomarker for future interventions and disease progression modelling.
Recent challenge, TADPOLE~\cite{tadpole_med} aimed at predicting future evolution of the AD patients, where the ventricles' volume was used as one of the main outcome measures.
Thus, we focus on the developmment of ventricles captured by brain MRI, and predicting their 3D segmentation evolution over a three-year period (Fig. \ref{fig:data_example}).

\subsection{Related work}

Addressing DPM with Deep Learning is an active area of research.
Existing methods can be broadly categorized into two approaches. In the first category are the methods that operate on tabular data summarizing volume-level quantitative biomarkers, such as the volume of relevant structures, which have been derived from the imaging data through manual or automated segmentation. Such \textit{image-level} methods typically employ Recurrent Neural Networks (RNNs) \cite{Wang2018,NGUYEN2020117203} to fit the temporal variation of the derived biomarker values. Methods in the second category utilize the raw imaging data directly for
spatio-temporal modelling of the disease, typically in the form of respective future lesion segmentation maps. We will refer to these methods as \textit{pixel-level} progression methods, and focus on them as this is the approach of our proposed method.
For modeling the GA progression,
Zhang et al. \cite{Zhang2019} proposed a pixel-level multi-scale deep Convolutional Neural Network \HB{(MS-DCNN)} for joint segmentation and prediction of GA. The proposed method takes preprocessed baseline en-face OCT projection and produces segmentation at the specified future visits. However, the method lacks the ability to interpolate the progression between the visits. In addition, the dataset used in experiments was small, consisting of only 29 patients. In a follow-up work, Zhang et al. \cite{ZHANG2021} proposed a Bi-LSTM-based prediction model and a UNet based refinement module. However, it requires two visits to make a prediction and the dataset used was still small, consisting of only 25 patients. Recently, Gigon et al.~\cite{taylor} proposed a pixel-level model that uses a Taylor series with learnable time derivatives to approximate patient-specific continuous GA growth. The method takes a set of retinal layer thickness maps as an input, which makes the method application-specific and limits the amount of information the model can learn from. As in previously mentioned works, the method was tested on a low amount of cases, precisely 20.

For the AD progression, Wang et al. \cite{Wang2018} proposed to utilize a image-level method with LSTM\cite{lstm} to forecast the AD stage at the upcoming visits using 78 features derived from demographics, patient history and other sources. In this method, the  prediction interval is provided as part of the input. Nguyen et al. \cite{NGUYEN2020117203} employed a similar approach utilizing MinimalRNN\cite{MinimalRNNTM}, where the authors linked each recursive update with one-month interval. The authors used the set of features provided as a part of TADPOLE\cite{tadpole_med} challenge. Among teams participating in TADPOLE\cite{tadpole_med}, Hill et al. \cite{tadpole_melba} as a part of team GlassFrog employed high-dimensional regression and disease state specific slope models and ensembling for ventricle volume forecasting. Ismail et al. \cite{tadpole_melba} as a part of BravoLab team proposed to use LSTM for making forecasts. Venkatraghavan et al.\cite{tadpole_melba} fitted a linear mixed-effects model to estimate ventricle volume expansion rate. The mentioned methods, however, 
don't exploit the information from the entire volume, since they rely on a set of input features, which were pre-defined by the human experts.

Several recent works utilized Recurrent Neural Networks (RNNs) \cite{Wang2018,NGUYEN2020117203}. Wang et al. \cite{Wang2018} proposed to use RNN trained on a set of biomarker values to predict the Alzheimer's Disease progression at the next visit. The proposed method supports irregular visit intervals by introducing an additional feature $\Delta t$ - denoting the time to the next visit. Nguyen et al. \cite{NGUYEN2020117203} trained a RNN on a set of \D{quantitative values} to predict a diagnosis and a set of Alzheimer's Disease biomarkers \Dc{biomarkes are already here} for every month up to six months in the future. The authors interpolated the missing visits in the data by generating them with the proposed model. This, however, doesn't handle variable time intervals.
A notable pixel-level method, was proposed by Ezhov et al. \cite{Ezhov2019NeuralPE}. The paper discussed the method to fit the parameters of the partial differential equation (PDE), that defines the brain tumor growth, to the real imaging data. Unfortunately, the solution is hardly generalizable outside the tumor growth tasks, as it would require knowing a form of PDE beforehand. In addition, the method was only tested on 2 real rat MRI scans. Another pixel-level method was proposed by Petersen et al. \cite{Petersen2019DeepPM}. The authors trained a deep learning model to learn tumor growth dynamics directly from the imaging data. The work focuses on modelling the distribution of the possible tumor appearances. However, as the authors mentioned, the method is neither suitable for predicting a single correct growth trajectory nor designed for it. Expanding the work on probabilistic disease modelling, Petersen et al.\cite{dgg} introduced a continuous progression model, based on UNet\cite{unet} and attention mechanism \cite{transformers}. As before, the model wasn't explicitly designed for predicting a single trajectory, but rather a distribution of possible outcomes. Additionally, the model was trained to utilize two or more images as prior information, which reduces the method's applicability in certain scenarios, such as screening for fast progressors from a single scan for enriching the clinical trials.

In this paper, we base our proposed pixel-level DPM method on NeuralODEs \cite{node}. The distinctive features of the NeuralODEs that make them especially suitable for DPM is that they are inherently continuous, which allows forecasting at any future time-point; they represent some underlying physical process, and, at the simplest level, the neural network learned by NeuralODEs correspond to the instantaneous change in the target variable. In addition, NeuralODEs allow for controlling the solution's properties and imposing domain-specific constraints on the learned function. This builds an ideal foundation for NeuralODEs as a disease progression model. 
With the mentioned advantages, NeuralODEs are steadily finding an application in learning disease progression. Hao et al. \cite{Hao} proposed a NeuralODE based diffusion model, conditioned on individual's connectivity, for Amyloid pathology progression, which is a primary pathological event in the Alzheimer's disease. De Brouwer et al. \cite{Brouwer2020LongitudinalMO} used their previously proposed GRU-ODE-Bayes model \cite{Brouwer2019GRUODEBayesCM} for prediction of disability progression in Multiple Sclerosis patients in a two-year interval using a set of biomarkers.

\subsection{Contribution}
All the mentioned methods for DPM share at least one of the drawbacks: they rely on hand-crafted features; they are inherently non-interpretable; they are designed for fixed forecasting intervals; they are time-variant systems, as they can produce different results for the same forecasting interval.
In contrast, we introduce a framework for pixel-level DPM based on NeuralODEs\cite{node} that addresses the above-mentioned drawbacks. To the best of our knowledge, this is the first work that incorporated NeuralODEs~\cite{node} for the disease progression and \HB{corresponding target anatomy} segmentation. The proposed method:
(1) extracts the relevant features directly from the image and provides progression estimate with corresponding segmentation masks at a future time-point for \HB{reliability and feasibility} assurance;
(2) learns from data with missing visits and variable visit intervals;
(3) interprets disease progression as autonomous ODE and introduces methods to incorporate prior domain-knowledge.
(4) introduces temporal Dice loss and highlights a successful application of NeuralODE \HB{as shown by our large-scale evaluation}

We trained the proposed method using two large-scale datasets and showed that the method learns \HB{accurate} segmentation maps and produces \HB{meaningful} progression estimates. We demonstrated the efficacy of the proposed method on a private longitudinal GA dataset with 967 scans of 100 patients, and two public AD datasets with 2537 and 286 scans of 633 and 143 patients each, far larger than in related works. We used one of the public dataset, TADPOLE D3-D4, as a hold-out \HB{test set} and compared our results with those of the participants in the TADPOLE challenge\cite{tadpole_melba}, achieving state-of-the-art prediction performance from the single baseline scan.

\section{Method}

\begin{figure*}[tb]
\centering
\includegraphics[width=0.8\textwidth]{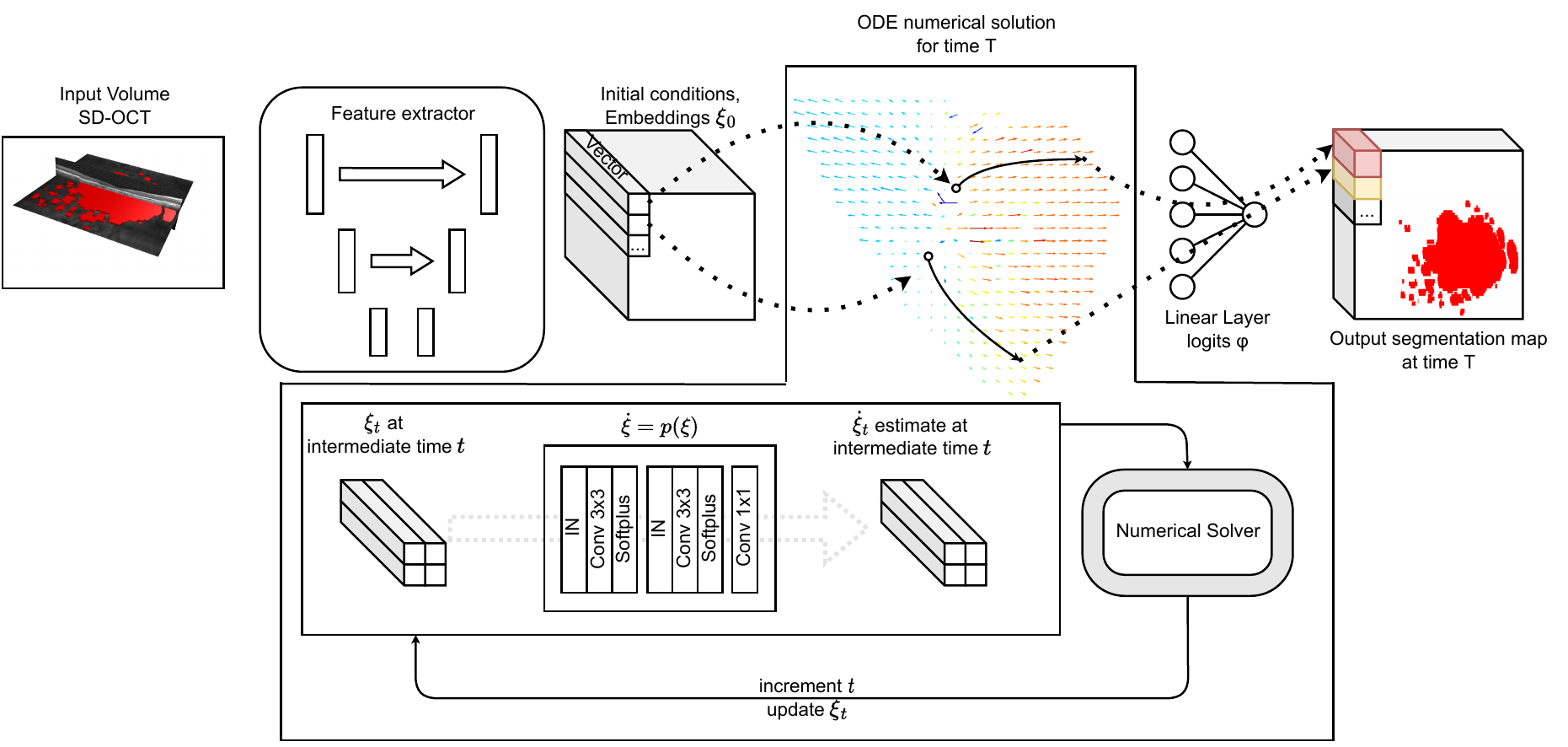}
\caption{Illustration of the proposed method. The input image at the baseline visit is first processed with a CNN to extract the feature embeddings $\xi_0$ for each pixel. The feature embedding $\xi_0$ acts as the initial condition for an ODE which evolves it  over time to form a trajectory for each pixel in the phase space. The solution trajectory is shown on a phase plot of Eq.~\ref{sys3} projected on 2D plane using PCA, where blue represents completely healthy pixels, and red represents pixels with GA. The computed solutions $\xi_T$ are projected with a  linear layer and sigmoid activation to obtain the output segmentation map $f_T$. The abstract procedure of numerical integration is presented at the bottom of the figure. The learnable function $p(\xi)$ representing time derivative of $xi$ takes form of a small CNN with 3 convolutional blocks. The computed embeddings $\xi_t$ at intermediate timepoint $t$ and function $p(\xi)$ forms an input to the black-box solver, which iteratively updates $\xi_t$.}

\label{fig:method}
\end{figure*}

We formulate the disease progression as an Initial Value Problem comprising an ordinary differential equation with an initial condition which specifies the value of the unknown function at baseline. The initial conditions take the form of the embedding vectors which are generated for each pixel of the baseline image using any segmentation architecture such as UNet\cite{unet}.
The ODE system defined on the embedding and logit space is numerically solved for the given time points in the future. The obtained logits corresponding to the future time point are then converted to the segmentation maps (Fig.~\ref{fig:method}) of the target anatomy. We utilize the method introduced in \cite{node} for training which requires a constant amount of memory independent of the solver's step size and integration time.



\subsection{Basic ODE model}
\label{method_secA}
 
We aim to learn a continuous spatio-temporal model of disease progression from a \HB{longitudinal} dataset comprising a series of $n$-dimensional images (3D in our case) for each subject, which are acquired across multiple visits at irregular time-intervals. The time-point $t \in \mathbb{R}^+$ of each visit is normalized to a scale of 1 year. Thus, $t=0.5$ and $t=1.5$ corresponds to the 6 and 18 month time-points respectively from the baseline visit at $t=0$. We represent the entire stack of the longitudinal volumetric data for each patient $i$ by a continuous function $x_i:\Omega\times\mathbb{R}^+\rightarrow\mathbb{R}$  such that $x_i(\omega; t$) is the intensity value at the spatial coordinate $\omega\in\Omega$ in the image acquired at time $t$. The entire image of the patient   acquired at time $t$ is denoted by $x(\cdot,t)$. The ground-truth segmentation masks for each image in the training set can be similarly represented by a continuous ground-truth function $g:\Omega\times\mathbb{R}^+\rightarrow [0,1]$, that corresponds to the state of the target biomarker. The continuous functions $x$ and $g$ are not directly observable but sampled at specific time-points $t$ corresponding to the visits during which the medical images were actually acquired. Our objective is to learn a continuous function $f:\Omega\times\mathbb{R}^+\rightarrow [0,1]$ to  approximate $g$ using a Deep Learning (DL) framework inspired from the NeuralODE \cite{node}. The proposed model attempts to approximate $f$ using only a single image $x(\cdot,0)$ of the baseline visit as input. The  estimated $f(\cdot,t)$ can then be sampled at arbitrary future time-points to obtain an estimate of the target anatomy segmentation in the future.

First, the image voxels of  $x_i(.,0)$ are projected onto the latent embedding in the form of convolutional feature maps $\xi_{0}$ using a CNN with learnable parameters $\Theta_\mathrm{CNN}$. In this work, the 3D-UNet \cite{unet3d} was used for MRI and the 3D$\rightarrow$2D PSC-UNet \cite{Lachinov2021ProjectiveSF} was employed for OCT. Subsequently, the evolution of $\xi_{t}$ over time is modeled as an Initial Value Problem (IVP) with the differential equation $\frac{\partial \xi}{\partial t}=p(\xi)$ and the initial condition $\xi=\xi_{0}$. The $p(\xi)$ is modeled using another CNN with three convolutional layers (see Fig \ref{fig:method} for details) with parameters $\Theta_\mathrm{ODE}$.
This IVP can be numerically solved to obtain the latent feature embedding for any arbitrary future target time-point $t$ and the corresponding evolving segmentation map is modeled as a linear transformation at each pixel (or voxel) of the 2D (or 3D) feature embedding. Thus, $f(\cdot;t)=\sigma(\phi _t)$, where $\phi_t=W^T \xi_t+b$ represents the unnormalized logits before applying the sigmoid activation to the linear transformation with the learnable parameters $W$ and $b$. During training, the backpropagation is performed as standard except for $\frac{\partial\xi_t}{\partial\Theta_\mathrm{ODE}}$ and $\frac{\partial\xi_t}{\partial\xi_0}$, whose computations are described in \cite{node}. The training and inference are summarized in the Algorithms \ref{alg_forward} and \ref{alg_backwards}, where ODE is one of the systems discussed below, ODESolve is a black box ODE solver, and $\Theta_\mathrm{ODE}$ are the learnable weights of the ODE.

\begin{algorithm}
\caption{Forward pass, $f(\cdot, t, x(\cdot,0), \Theta_{CNN}, \Theta_{ODE}, W)$}
\label{alg_forward}
\begin{algorithmic}
\REQUIRE Baseline image $g(\cdot,0)$, target time $t$, parameters $\Theta$ and $W$
\begin{itemize}
\item Compute embeddings $\xi_0~= \mathrm{CNN}(x(\cdot,0),\Theta_\mathrm{CNN})$,

\item Solve ODE, $\xi_t = \text{ODESolve}(\xi_0,\text{ODE},t,\Theta_\mathrm{ODE})$, $\phi_t = W^T\xi_t+b$

\item Compute the final segmentation maps with sigmoid, $\sigma(\phi_t)$
\end{itemize}
\end{algorithmic}
\end{algorithm}

\begin{algorithm}
\caption{Backward pass}
\label{alg_backwards}
\begin{algorithmic}
\REQUIRE Baseline image $x(\cdot,0)$, annotations $g(\cdot,0)$ and $g(\cdot,t)$, target time $t$, parameters $\Theta_{CNN}$, $\Theta_{ODE}$ and $W$
\begin{itemize}
\item Compute $f(\cdot, t, x(\cdot,0), \Theta_{CNN}, \Theta_{ODE}, W)$ with forward pass,
\item Compute loss $L(f,g)$,
\item Compute $\frac{\partial L}{\partial \xi_t}$ and $\frac{\partial L}{\partial W}$ as regular, compute $\frac{\partial\xi_t}{\partial\xi_0}$ and $\frac{\partial\xi_t}{\partial\Theta_\mathrm{ODE}}$ as in \cite{node}, compute $\frac{\partial\xi_0}{\partial\Theta_\mathrm{CNN}}$ as regular,
\item perform an update of $\Theta_\mathrm{ODE}$, $\Theta_\mathrm{CNN}$ and $W$.
\end{itemize}
\end{algorithmic}
\end{algorithm}

\subsection{Variations of the proposed ODE}

The proposed IVP based formulation for DPM gives us a higher degree of control over the evolution of the disease. It allows the incorporation of additional \HB{domain-specific} anatomical constraints by employing different systems of ODE in the IVP. In certain medical conditions, once a pathology occurs at a spatial location, the damaged tissue can never be recovered. In such cases, for example in GA, the target biomarker can only increase in size over time.  
We first summarize the basic IVP formulation \HB{without constraints} as described previously. 
Next, we propose two modifications to the ODE system to enforce the \textit{non-decreasing} constraints on the size of the target anatomy progression. 

\begin{itemize}
    \item \textit{No constraints:}
    This simple approach does not impose any constraints on the way the biomarker segmentation evolves over time. In this case $\phi_t$ is obtained as a linear transformation of the embedding feature space $\xi$ that evolves over time as defined by the ODE  
    \begin{align}
    \label{sys1}
    \dot{\xi} = p(\xi),
    \end{align}
    where the function $p(\xi)$ is modeled using a learnable CNN to predict $\dot{\xi}$, which is a shorthand notation for the time derivative $\frac{\partial \xi}{\partial t}$. The ODE is assumed to be autonomous or time-invariant implying that the function $p(\xi)$ is independent of time, ie., the CNN neither takes $t$ as input, nor does its network parameters change with time.

    \item \textit{Constraints on the embeddings: }
    The \HB{non-decreasing} target anatomy constraint can be incorporated into our ODE system by enforcing a non-negativity constraint on the derivative of $\phi$ with respect to time, ie., $\dot{\phi}\ge0$.
    Given the linear dependency of the logits $\phi = W^T\xi+b$, we need $\dot{\phi} = W^T\dot{\xi}=W^Tp(\xi)$ to be non-negative. Considering the unit vector $\hat{W}=\frac{W}{||W||}$, we can decompose $p(\xi)$ into orthogonal and collinear components $p(\xi)= \hat{W^T}p(\xi)\hat{W} + (p(\xi) - \hat{W^T}p(\xi)\hat{W}$).
    We explicitly constrain the projection of $p(\xi)$ on $\hat{W}$ in the collinear term to be non-negative by clipping its negative values to 0 using a rectified unit function denoted by $|\cdot|^+$. Thus, the proposed ODE becomes
    \begin{align}
    \label{sys3}
    \dot{\xi}=|\hat{W}^T p(\xi)|^+ \hat{W} + p(\xi) - \hat{W}^T p(\xi) \hat{W}.   
    \end{align}
    

    
    \item \textit{Constraints on the logits:}
    In contrast to the above, where $\dot\phi$ and $\dot\xi$ had a linear dependency, we want to decouple these two variables and introduce non-linear dependency between them. Thus, we introduce the time derivative of $\phi$ as a learnable function $q(\xi,\phi)$, and apply explicit non-negative constraint by the means of rectified unit function.
    In this case, we introduce the following system:
    \begin{align}
    \label{sys2}    
    \begin{cases}
    \dot{\xi}=p(\xi,\phi),\\
    \dot{\phi}=|q(\xi,\phi)|^+,
    \end{cases}
    \end{align}
    where we used \textit{softplus} as rectified unit function $|\cdot|^+$.
\end{itemize}

In the implementation of $p(\xi)$, we used two consecutive residual blocks (Fig.~\ref{fig:method}) consisting of instance normalization~\cite{instancenorm}, convolution with filter size $3\times3$ and softplus activation followed by $1\times1$ convolution (for 2D, with natural extension to 3D case). For implementing Eq.~\ref{sys2}, we concatenate $\xi$ and $\phi$ and implement $q$ identically to $p$.


\subsection{Loss function}
The proposed method is trained by obtaining $f(\cdot, t)$ from the ODE at specific discrete time points $\lbrace{0 \le t \le T\rbrace}$, corresponding to the actual patient visits for which the GT $g(\cdot,t)$ are available. The total loss at a subject-level is then be defined as the mean of the \HB{segmentation losses across all visits}.
The binary cross-entropy could be used at each time step:
\begin{flalign}
\label{loss3}
\bar D(f,g) = - \sum_{t=0}^T\sum_{\omega\in \Omega}{g(\omega,t)\log(f(\omega,t))} +\nonumber\\
(1-g(\omega,t))\log(1 - f(\omega,t))
\end{flalign}
Another alternative is the soft Dice loss \cite{vnet}:
\begin{flalign}
\label{loss2}
\bar D(f,g) = 1 - \frac{1}{T}\sum_{t=0}^{T}\frac{2\sum_{\omega\in \Omega}{f(\omega,t)g(\omega,t)}}{\sum_{\omega\in \Omega}{f^2(\omega,t)} + \sum_{\omega\in \Omega}{g^2(\omega,t)}}
\end{flalign}

Below, we approach the problem of the temporal loss using functional analysis and derive the temporal Dice loss function. We define $f,g:\Omega\times\mathbb{R}^+\rightarrow[0;1]$, depending on time, and assume that $f$ and $g$ are continuous almost everywhere on $\Omega\times\mathbb{R}^+$, and that both integrals of $f$, $g$ and their squares are finite. In turn, the space of such square integrable functions $F= \{f: \Omega\times\mathbb{R}^+\rightarrow [0;1]\}$ with inner product defined as $(f,g)=\int_{\Omega\times\mathbb{R}^+}{f(\omega,t)g(\omega,t)d\omega dt}$ is a Hilbert space.

We want our method to learn the function $f$ by minimizing the respective squared distance $\lVert f-g\rVert^2$.
$$
\lVert f-g\rVert^2 = \lVert f\rVert^2 + \lVert g\rVert^2 - 2(f,g) \leq \lVert f\rVert^2 + \lVert g\rVert^2,
$$ due to non-negativity of $(f,g)$.

The minimum of $\lVert f-g\rVert^2$ is achieved at $f \sim g$: $f$ and $g$ being from the same equivalence class. Introducing a surrogate function $D(f,g)$, we limit it's the upper bound to~1: $D(f,g) = \frac{\lVert f-g\rVert^2}{\lVert f\rVert^2 + \lVert g\rVert^2} $.
We can show that $D(f,g)$ has a minimum at $f \sim g$. As a result, by finding the minimum of $D(f,g)$ we also minimize the distance $\lVert f-g\rVert^2$.
Next, we discretize the space $\Omega$ and time $T$, thus getting the definition of Soft Dice loss \cite{vnet} from $D(f,g)$ with temporal objective:
\begin{flalign}
\label{loss1}
&\bar D(f,g) = \nonumber\\ 
&1 - \frac{2\sum_{t=0}^T\sum_{\omega\in \Omega}{f(\omega,t)g(\omega,t)}}{\sum_{t=0}^T\sum_{\omega\in \Omega}{f^2(\omega,t)} + \sum_{t=0}^T\sum_{\omega\in \Omega}{g^2(\omega,t)}}.
\end{flalign}
Effectively, this implies that we should compute Dice loss not for individual visits independently, but for the entire stack of visits. In the experiments, as part of ablation, we compare the temporal Dice formulation with the other two loss functions and study their influence on the performance.

\section{Evaluation}

\subsection{Data}
Table~\ref{tab_datasets} provides an overview of the three datasets used to develop and evaluate the proposed method. 
The private MUV-GA dataset is used to evaluate the performance on the task of predicting the segmentation maps for GA at future time-points from a retinal OCT of the first baseline visit. The two public datasets TADPOLE \cite{tadpole_med} and ADNI1 are employed for the task of predicting the changes in the ventricle volume over time, which is an important biomarker to track the AD progression. 



\renewcommand{\arraystretch}{1.0}
\begin{table}[th]
\centering
\caption{An overview of the longitudinal Datasets.}
\label{tab_datasets}
\resizebox{0.46\textwidth}{!}{

\begin{tabular}{lcccccc}
\toprule
\thead{Dataset} & \thead{Scan} & \thead{\#Patients} & \thead{Total\\\#scans} & \thead{ \#Scans\\per patient} & \thead{Avg. time\\span (days)} \\
\midrule
MUV-GA & \thead{OCT} & 100 & 967 & 5.95  & 970 \\
\midrule
ADNI & MRI & 633 & 2537 & 4 & 704 \\
\midrule
\thead{TADPOLE\\D3-D4}
 & MRI & 143 & 286 & 2 & 1082 \\
\bottomrule

\end{tabular}}
\end{table}

\subsubsection{MUV-GA} is a longitudinal in-house dataset from the \HB{Department of Ophthalmology} at Medical University of Vienna~\cite{Bui2021} which originated from an observational study of natural GA progression. It consists of 3D OCT scans of eyes of patients older than 50 years, diagnosed with GA, imaged with Spectralis (Heidelberg Engineering) device. Study visits occurred every 3 months. The patients were diagnosed with GA secondary to AMD, and were excluded from the study when other confounding pathologies were found. 
It consists of 3D retinal OCT scans of subjects  older than 50 years who have been diagnosed with GA. Longitudinal scans were acquired for each subject at 3 month visit intervals using the Spectralis (Heidelberg Engineering) device.
The Ground Truth pixel-level annotations of GA lesion were
performed manually by a clinical expert at all available time-points.
The dataset features both missing visits and irregular time intervals between visits. While the OCT scans are 3D, the reference standard for the GA segmentation is available in 2D, corresponding to the en-face lesion as depicted in Fig.~\ref{fig:data_example}. Thus, the CNN architectures that can be employed for this task are limited to 3D$\rightarrow$2D networks, such as \cite{Lachinov2021ProjectiveSF}, which reduce the dimensionality of the output predictions.

\paragraph{Preprocesing}
First, we resample the scans to have 5.671$\mu\text{m}$ spacing across the BScans. Then, we register the follow-up visits. To control and fix the alignment of the scans and corresponding annotations between visits, we first perform retinal blood vessel segmentation~\cite{Lachinov2021ProjectiveSF}, and then use phase correlation algorithm~\cite{4767966} on the blood vessels to estimate the shift between the given image and the baseline scan. After the registration of the follow-up visits is done, we run layer segmentation algorithm\cite{iowa} and flatten the retinal appearance according to the outer RPE boundary. Flattening helps the methods to focus on retinal layer appearance and structure, rather than spatial position and inclination.

\subsubsection{ADNI} 
ADNI1\footnote{adni.loni.usc.edu} is an open-access dataset - part of the initiative launched in 2003  as  a  public-private  partnership,  led  by  Principal  Investigator    Michael    W.    Weiner,  MD. The dataset was created for measuring the progression of mild cognitive impairment and early Alzheimer's Disease. The primary goal of ADNI has been to test whether serial MRI,  positron  emission  tomography  (PET),  other  biological  markers,  and  clinical  and neuropsychological  assessment  can  be  combined  to  measure  the  progression  of  mild cognitive impairment (MCI) and early Alzheimer’s disease. 

TADPOLE\cite{tadpole_med} is a challenge guided towards the accurate prediction of AD progression. It aims to predict AD progression through estimation of a number of biomarker values. In this work, we were specifically interested in the imaging biomarkers, thus we focused on the prediction of the ventricle volume. The TADPOLE challenge has a series of datasets, derived from ADNI database.
In this study, we use only \textbf{D3 and D4} datasets, designed for making predictions from just a single baseline scan, where D3 contains \HB{a single visit for a patient}, and D4 plays a role of the test set with the follow-up measurements.
We use TADPOLE data exclusively \HB{as a hold-out test-set} and train on \textbf{ADNI1} data.

As the proposed method requires imaging data to train, we query T1 MRI scans of the patients participating in ADNI1 and having at least 2 visits. Due to the absence of the manual ground truth ventricle annotation, we obtain pseudo-label segmentations automatically using FastSurfer~\cite{fastsurfer} at all available time points following the protocols used in the TADPOLE challenge. We then trained the baseline methods and the proposed method using such pseudo-label segmentations, and test using TADPOLE D3-D4 data. Both the input MRI images and the corresponding segmentation labels are in 3D. Similar to MUV-GA, TADPOLE has missing visits and irregular time intervals between visits.

\paragraph{Preprocessing}
The MRI scans were first imported to the FreeSurfer, normalized\cite{freesurfer1} and skull-stripped\cite{freesurfer2}. Then, FastSurfer \cite{fastsurfer} was used to process the scans and segment the brain structures. Follow-up visits of the same patients were aligned using longitudinal stream of FreeSurfer \cite{freesurfer3}. Since we explicitly focus on the evolution of the ventricle volume, we discard all other labels. The preprocessed scans have the size of $256\times 256\times 256$. To reduce the memory burden for the storage and data processing, we additionally crop the cube of size $128\times 128\times 128$. We center the cropped region on the ventricles of the baseline visit. The follow-up scans are cropped with the region from the baseline.

\subsection{Experiments}
\subsubsection{Data split}
To assess the performance of the proposed method, we run a series of experiment for each dataset. In our experiments, we utilize 5-Fold Cross Validation (CV) and split \textbf{VSC GA} and \textbf{ADNI} datasets into five groups by patient id, where four folds are used as a train-val dataset, and one fold is used as a test dataset.
We use \textbf{TADPOLE D3-D4} exclusively as a hold-out test set, and those scans\HBc{but also patients?}\Dc{these are +- same patients, adni1 are visits from \~2000, tadpole d3-d4 are modern 2018-2020} don't overlap with \textbf{ADNI1}.

\subsubsection{Baselines}
Since the datasets represent different tasks, 
we compare the proposed method with different baselines depending on the application. 

\paragraph{Oracle} Each of the datasets has a corresponding \textit{Oracle} model as an estimate of the upper bound that a predictive model can achieve. To obtain an \textit{Oracle} model, we directly segmented both the baseline and the endpoint scan. For this, we trained PSC-UNet~\cite{Lachinov2021ProjectiveSF} for automated GA segmentation, and UNet3D \cite{unet3d} to perform an automated segmentation of the brain ventricles.

\paragraph{Geographic atrophy} We compare the proposed method against the approach introduced in Zhang et al. referred as MS-DCNN\cite{Zhang2019}, Nguyen et al. \cite{NGUYEN2020117203}, Gigon et al.\cite{taylor} and the \textit{Oracle} model. The nature of the task of predicting growth on 2D en-face masks from 3D volumetric data (3D $\longrightarrow$ 2D) poses an additional challenge and limits a set of the \HB{baseline} methods that can be utilized. The performance of progression prediction is evaluated at the follow-up visit, one year from the baseline scan.

\paragraph{ADNI \& TADPOLE}
For ADNI data, we compare the proposed method with Nguyen et al. \cite{NGUYEN2020117203}, Petersen et al. Deep Glioma Growth (DGG)\cite{dgg} and \textit{Oracle} model. The prediction performance is evaluated at the 3-year follow-up visit.
For TADPOLE D3-D4, we additionally include the methods from TADPOLE challenge in comparison, namely \textit{GlassFrog}, \textit{BravoLab}, \textit{EMC1} \cite{tadpole_melba}.

\subsubsection{Ablation experiments}

We ran a series of ablations experiments, to investigate:

\paragraph{Constraints in the ODE system}
we train the proposed method with three different ODE systems, one introducing no constraints (Eq.~\ref{sys1}), one constraining the  embedding space (Eq.~\ref{sys3}), and one constraining the logits space (Eq. \ref{sys2}). For this ablation experiment, we employed only \textbf{MUV-GA} dataset, because GA can only grow in size as the atrophic tissue cannot recover, \HB{which is not necessarily the case with brain ventricles.} As a loss, we employed the function described in Eq. \ref{loss1}.

\paragraph{Loss functions} for learning temporal segmentation objective we investigate the performance of the loss function derived from functional distance (Eq. \ref{loss1}), compared to the alternative standard loss functions (Eq. \ref{loss3}, and \ref{loss2}). We perform a 5-fold cross validation experiments using the proposed method on both \textbf{MUV-GA} and \textbf{ADNI} datasets, and further apply the models trained on \textbf{ADNI} to \textbf{TADPOLE} hold-out test set. For \textbf{MUV-GA} dataset, we employed the ODE system with constrains defined by Eq. \ref{sys3}. For \textbf{ADNI} data, we used ODE system described by Eq. \ref{sys1}.






\subsubsection{Evaluation metrics}
For cross-validation experiments on \textbf{MUV-GA} and \textbf{ADNI} datasets, we report: Dice index of baseline and predicted visit, Hausdorff distance of the baseline and the last visit, coefficient of determination $R^2$ and coefficient of correlation Pearson's $r$ of \textit{the change in area or volume}. For \textbf{MUV-GA}, we additionally report the mean squared error (MSE) of the change in square root transformed area in mm. For \textbf{ADNI} and \textbf{TADPOLE}, we report the metrics utilized in TADPOLE\cite{tadpole_med} challenge. Namely, MAE, WES, calculated as percentage of intracranial volume, and CPA:
$$MAE = \frac{1}{N}\sum_{i=1}^N{\frac{1}{ICV_i}|V_i - \bar V_i| },$$
$$WES = \frac{\sum_{i=1}^N{\frac{1}{\bar V_i^{0.75} - \bar V_i^{0.25}}|V_i - \bar V_i|}}{\sum_{i=1}^N{\frac{ICV_i}{\bar V_i^{0.75} - \bar V_i^{0.25}}}},$$
$$CPA=|0.5 - \frac{1}{N}\sum_{i=1}^N{1_{[\bar V_i^{0.25};\bar V_i^{0.75}]}(V_i)}|,$$
where $V$ and $\bar V$ are predicted ventricle volumes, $\bar V_i^{0.25}$ and $\bar V_i^{0.75}$ are 25\% and 75\% percentile of the predicted volume of an individual patient, together forming 50\% CI, $ICV$ is the intracranial volume, and 1 is an indicator function.

\begin{table*}[h]
\centering
\caption{Cross-validation experiments on MUV-GA dataset. The metrics were computed with the respect to the baseline and the 1-year visits. In contrast to the other methods, the Oracle model does not perform a future prediction but employs the Scan at the 1 year visit to highlight an upper limit of the performance that can be achieved. }
\label{tab_vsc_ga_cv}
\begin{tabular}{l|cc|cc|ccc}
\hline
Method & Dice$_0$ & Dice$_t$ & HD$^{95}_0$ & HD$^{95}_t$ & $R^2$ & $r$ & MAE (in mm)\\
\hline
\HB{Oracle:} PSC-UNet\cite{Lachinov2021ProjectiveSF}& $0.82\pm0.15$ & $0.87\pm0.13$ & $0.16\pm0.21$ & $0.10\pm28$ & 0.73 & 0.86 & $0.46\pm0.59$  \\
\hline
Zhang et al. \cite{Zhang2019} & $0.62\pm0.25$ & $0.70\pm0.21$ & $1.50\pm1.00$ & $1.05\pm0.74$ & -0.18 & 0.52 & $1.19\pm0.96$\\
Zhang et al. 3D \cite{Zhang2019} & $0.76\pm0.21$ & $0.81\pm0.14$ & $0.78\pm1.12$ & $0.47\pm0.69$ & -0.29 & 0.38 & $1.15\pm1.10$\\
Nguyen et al. \cite{NGUYEN2020117203} & $\mathbf{0.80\pm0.17}$ & $\mathbf{0.84\pm0.11}$ & $\mathbf{0.30\pm0.54}$ & $\mathbf{0.19\pm0.29}$ & 0.28 & 0.66 & $0.89\pm0.80$\\
Gigon et al. \cite{taylor} & $0.76\pm0.22$ & $0.79\pm0.18$ & $0.57\pm0.89$ & $0.78\pm1.12$ & 0.36 & 0.61 & $0.85\pm0.75$\\
\hline
\textbf{Proposed} & $\mathbf{0.80\pm0.19}$ & $\mathbf{0.84\pm0.13}$ & $\mathbf{0.30\pm0.54}$ & $0.21\pm0.30$ & \textbf{0.59} & \textbf{0.78} & $\mathbf{0.64 \pm 0.63}$\\
\hline
\end{tabular}
\end{table*}

\begin{table*}[tb]

\centering
\caption{Cross-validation experiments on ADNI dataset. The metrics were computed with the respect to the baseline and the 3-year visits. The Oracle model highlights the predictive ceiling that can be achieved.}
\label{tab_adni_cv}
\begin{tabular}{c|ll|ll|ll|lll}
\hline
Method & Dice$_0$ & Dice$_t$ & HD$^{95}_0$ & HD$^{95}_t$ & $R^2$ & $r$ & MAE, \% ICV & WES & CPA\\
\hline
\thead{Oracle: 3D-UNet~\cite{unet3d}} & $0.97\pm0.02$ & $0.97\pm0.2$ & $1.08\pm4.20$ & $1.28\pm4.32$ & 0.98 & 0.99 & $0.06\pm0.10$ & 0.05 & 0.46 \\
\hline
\hline
Nguyen et al.\cite{NGUYEN2020117203}& $0.96\pm0.02$ & $0.93\pm0.04$ & $1.46\pm4.31$ & $1.90\pm5.07$ & 0.32 & 0.58 & $0.24\pm0.20$ & 0.23 & \textbf{0.04} \\
\thead{Petersen et al.\cite{dgg},\\mean sample} & $0.95\pm0.02$ & $0.91\pm0.04$ & $\mathbf{1.10\pm4.20}$ & $\mathbf{1.38\pm4.40}$ & -0.11 & 0.34 & $0.24\pm0.26$ & 0.19 & 0.25 \\
\thead{Petersen et al.\cite{dgg},\\best sample} & $0.95\pm0.02$ & $0.91\pm0.04$ & $\mathbf{1.10\pm4.20}$ & $\mathbf{1.38\pm4.30}$ & -0.09 & 0.35 & $0.24\pm0.26$ & \textbf{0.18} & 0.25 \\
\hline
\textbf{Proposed} & $\mathbf{0.97\pm0.02}$ & $\mathbf{0.94\pm0.03}$ & $1.39\pm4.21$ & $1.63\pm4.63$ & \textbf{0.36} & \textbf{0.61} & $\mathbf{0.21\pm0.19}$ & 0.20 & 0.08 \\
\hline
\end{tabular}
\end{table*}

\subsection{Training details}

Each dataset was split into five folds. For each method, we performed a brief hyperparameter optimization based on the additional validation subsplit of the training folds. We chose between ADAM and SGD optimizers and a number of learning rates. For the development We used PyTorch framework, and we trained all the models using NVidia 2080Ti GPU. Training of each model took less than 24h.

\paragraph{Proposed method} We used the function defined by Eq.~\ref{loss1} as a loss function. For solving the ODE system, we used RK4 scheme with step size $h=\frac{1}{12}$ equivalent to a one-month interval. For optimization, we used ADAM optimizer with default parameters for all 3 variants of the proposed method. For NeuralODE part, we used the code provided by Chen et al.\footnote{https://github.com/rtqichen/torchdiffeq} 
During training the proposed method takes a constant amount of GPU memory, independent of the time-interval over which the ODE is evolved by using the adjoint method  \cite{node} for backpropagation. This allowed our method to be trained over large time intervals with limited GPU memory.

\paragraph{\HB{Baseline methods}}
For \HB{MS-DCNN}~\cite{Zhang2019}, we removed dropout and added InstanceNorm \cite{instancenorm}, which greatly improved training stability and performance. We also extended the method of Zhang et al. to handle 3D OCT volumes instead of en-face projections, it is referred to as MS-DCNN 3D.
We extended the approach of Nguyen et al. \cite{NGUYEN2020117203} to handle the imaging data instead of tabular features, introducing convolutional filters into the recurrent blocks. The work of Gigon et al. \cite{taylor} was implemented as in the paper, however, we used \cite{iowa} for layer segmentation. For implementation of the Deep Glioma Growth\cite{dgg}, we used code provided by Petersen et al.\footnote{https://github.com/MIC-DKFZ/image-time-series}, with the exception that only baseline image and its segmentation was used for generating the prior.

\section{Results}

\begin{table}[t]
\centering
\caption{Results for TADPOLE D4 test set, evaluated using a single baseline scan.\HBc{Why Dice 0 and not Dice t?}\Dc{no access to the MRI at t}\HBc{What do broken lines denote?}\Dc{spacers between the tadpole methods. Teamnames are multiline, it's hard to navigate the table without the lines.}}
\label{tab_tadpole_test}
\begin{tabular}{c|ll|lll}
\hline
Method & Dice$_0$ & HD$^{95}_0$ & \makecell{MAE,\\ \% ICV} & WES & CPA\\
\hline
Nguyen \cite{NGUYEN2020117203} & $\mathbf{0.96\pm0.01}$ & $1.07\pm0.46$ & $0.46$ & 0.39 & \textbf{0.02}\\
\thead{Petersen \cite{dgg}.\\mean sample} & $0.95\pm0.01$ & $\mathbf{1.0\pm0.4}$ & $0.56$ & 0.4 & 0.2\\
\textbf{Proposed} & $\mathbf{0.96\pm0.01}$ & $1.05\pm0.44$ & $\mathbf{0.44}$ & \textbf{0.35} & 0.08\\
\hline
\makecell{GlassFrog-\\LCMEM\\-HDR\cite{tadpole_melba}} & - & - & 0.48 & 0.38 & 0.24\\
\hdashline
BravoLab\cite{tadpole_melba} & - & - & 0.64 & 0.64 & 0.42\\
\hdashline
\makecell{GlassFrog-\\Average\cite{tadpole_melba}} & - & - & 0.68 & 0.55 & 0.24\\
\hdashline
\makecell{EMC1-\\Std\cite{tadpole_melba}} & - & - & 0.80 & 0.62 & 0.48\\
\hline
\end{tabular}
\end{table}

\begin{table}[t]
\centering
\caption{Ablation experiments of explicit domain-specific constraints. Eq.~\ref{sys1} corresponds to the system without constraints. To encode that the lesion only grows over time, Eq. \ref{sys3} uses explicit constrains in the embedding space, while Eq.~\ref{sys2} applies the constraints on the logits.}
\label{tab_ablations_eq}
\begin{tabular}{c|ll|lll}
\hline
ODE & Dice$_0$ & Dice$_t$ & R$^2$ & r & MAE, mm\\
\hline
Eq. (\ref{sys1}) & $0.79\pm0.19$ & $0.83\pm0.15$ & 0.44 & 0.72 & $0.78\pm0.69$\\
Eq. (\ref{sys3}) & $\mathbf{0.8\pm0.19}$ & $\mathbf{0.84\pm0.13}$ & $\mathbf{0.59}$ & \textbf{0.78} & $\mathbf{0.64\pm0.63}$\\
Eq. (\ref{sys2})& $\mathbf{0.8\pm0.19}$ & $\mathbf{0.84\pm0.14}$ & $0.56$ & 0.76 & $0.68\pm0.63$\\
\hline
\end{tabular}
\end{table}

\begin{table*}[t]
\centering
\caption{Ablation experiments of the proposed method trained with different loss functions. Eq. \ref{loss3} represents binary cross entropy loss, Eq. \ref{loss2} reflects a soft Dice loss applied on each individual scans and then averaged, Eq. \ref{loss1} represents the Dice loss described in the paper, considering all scans as a \HB{single 4D volume}.}
\label{tab_ablations_loss}
\begin{tabular}{c|ll|lll|ll|ll|lll}
\hline
Dataset & \multicolumn{5}{c|}{MUV-GA, 5-Fold CV} & \multicolumn{4}{c|}{ADNI, 5-Fold CV} & \multicolumn{3}{c}{TADPOLE D4} \\
\hline
Loss & Dice$_0$ & Dice$_t$ & R$^2$ & r & MAE, mm & Dice$_0$ & Dice$_t$ & R$^2$ & r & MAE, \% ICV & WES & CPA\\
\hline
Eq. (\ref{loss3}) & $0.73\pm0.22$ & $0.77\pm0.18$ & $0.46$ & 0.68 & $0.71\pm0.74$ & $0.97\pm0.02$ & $0.94\pm0.03$ & $0.36$ & $\mathbf{0.6}$ & $0.46\pm0.8$ & $0.36$ & $0.12$ \\
Eq. (\ref{loss2})& $0.78\pm0.18$ & $0.83\pm0.13$ & $0.58$ & 0.76 & $0.67\pm0.61$ & $0.97\pm0.02$ & $0.94\pm0.03$ & $0.36$ & $\mathbf{0.6}$ & $0.45\pm0.82$ & $0.36$ & $\mathbf{0.05}$\\
Eq. (\ref{loss1}) & $\mathbf{0.8\pm0.19}$ & $\mathbf{0.84\pm0.13}$ & $\mathbf{0.59}$ & \textbf{0.78} & $\mathbf{0.64\pm0.63}$ & $0.97\pm0.02$ & $0.94\pm0.03$ & $0.36$ & $0.61$ & $\mathbf{0.44\pm0.83}$ & $\mathbf{0.35}$ & $0.08$\\
\hline
\end{tabular}
\end{table*}

\subsection{Progression of Geographic Atrophy} 
A qualitative result of our method is shown in Fig.~\ref{fig:data_example}, and the quantitative cross-validation results are provided in Table~\ref{tab_vsc_ga_cv}.
The top row represents the segmentation performed with \textit{Oracle} PSC-UNet\cite{Lachinov2021ProjectiveSF} using baseline and first year images.
The proposed method clearly outperformed the baseline methods on all the metrics with the exception of Nguyen et al. \cite{NGUYEN2020117203}, where they both achieved a mean Dice score of 0.8 and 0.84 for predicting the atrophy segmentation at baseline and 1-year visits, respectively.
However, the proposed method performed better in estimating the atrophy growth, achieving $R^2=0.59$, $r=0.78$,  and MAE $=0.64$ mm, which is very close to the performance of the \textit{Oracle} model. 
The similar performance of  the proposed method and the RNN based solution of \cite{NGUYEN2020117203} may be attributed to the fact that both belong to the same family with the RNN model being a discrete approximation of the NeuralODE. The RNN can be represented as NeuralODE with a less sophisticated explicit Euler method for numerical integration at discrete time-steps.

The recently introduced method of Gigon et al. \cite{taylor}, exploring the idea of Taylor series decomposition, performs a bit worse both in segmentation and prediction. The method of Zhang et al. \cite{Zhang2019} failed in quantitative prediction of the lesion change, at the same time showing moderate performance in terms of segmentation.

\subsection{Progression of Alzheimer's Disease}
A qualitative result of our method for predicting the ventricle segmentation in brain MRI at future time-points is depicted in Fig.~\ref{fig:data_example}.
All the models achieved a high segmentation performance of the baseline visits, with Dice score ranging from 0.95 of Petersen et al. \cite{dgg} to 0.97 of the proposed method (Table \ref{tab_adni_cv}).
It's important to note that we used automatically generated annotations by FastSurfer, which potentially might not fully reflect the complexity of the ventricle anatomy and might be easy to learn by automated methods. 
The proposed method achieved the best performance in predicting segmentation at a 3-year visit in terms of Dice, $R^2$, $r$ and MAE. The \HB{other} methods demonstrated similarly good results in baseline\HBc{unfortunate that you have baseline as a scan and baseline as a method} and 3-year HD$^{95}$, WES and CPA.
As Petersen et al. claim, the Continuous-Time Deep Glioma Growth model wasn't designed for precise prediction of the changes, but rather for their probabilistic description. However, the Deep Glioma Growth model achieved good results in predicting the changes, with mean prediction being close to the best sample from the predicted segmentation masks distribution w.r.t 3-year Dice score.
 
The TADPOLE \HB{hold-out} test set consists of on average longer forecasting intervals, testing the extrapolation capabilities of the methods. Quantitatively (Table~\ref{tab_tadpole_test}), the proposed and baseline models demonstrated good generalization in baseline visit segmentation, achieving Dice score of 0.95 or above and showing a mean HD$^{95}$ smaller than 1.07 mm.
Importantly, the proposed method, as in the cross-validation experiments, excelled in accurate disease evolution prediction, surpassing the challenge contenders in all metrics. It clearly outperformed the  TADPOLE participants, in the prediction of the future ventricle size, achieving 0.44 MAE, 0.35 WES, and demonstrating a competitive 0.02 CPA. The method of Petersen et al. \cite{dgg} showed competitive predictive performance, where we sampled the mean prediction, since we were interested in a single prediction output.
Overall, all of the imaging-based method demonstrated competitive results, with the proposed method surpassing the challenge contenders in all metrics.

\subsection{Ablation experiments.}
To investigate the effect of the explicit constraint enforcement as a part of the ODE system, we compared three different ODE systems defined by Eq.~\ref{sys1}-\ref{sys2} (Table~\ref{tab_ablations_eq}) on the MUV-GA dataset. 
The effect of the constrained ODE system was slightly pronounced on the segmentation performance of the baseline and 1-year scans as both ODE systems with constraints (Eq. \ref{sys3} and \ref{sys2}) performed slightly better than the unconstrained system. However, for the atrophy growth estimation, the effect of the constraints was more pronounced, where the system with the constraints in the embedding space (Eq.~\ref{sys3}) performed clearly the best.

The effect of different loss functions on the predictive performance is reported in Table~\ref{tab_ablations_loss}. 
In all experiments, binary cross-entropy function performed the worst, except in ADNI cross-validation, where all \HB{loss functions} performed equally well. This can be attributed to the nature of the dataset's reference standard, which was produced by an automated segmentation method\HBc{but this was also the case in TADPOLE as well?}\Dc{partly, tadpole has no segmentation}. Both Dice-based losses performed well, however, the loss we derived from functional distance (Eq.~\ref{loss1}), performed better in GA.
 
\section{Conclusion}



In this work, we introduced a DPM to predict the future progression of the morphological changes in pathologies (such as GA in retinal OCT) and clinically relevant anatomical structures (such as the ventricles in brain MRI) from a single scan of the baseline visit. Our method produces the segmentation maps for the desired object of interest for varying future time-points which is clinically useful in the early detection of the fast progressors for personalized treatment and the recruitment of patients in clinical trials. 

In contrast to the existing methods, the proposed method employs a continuous time-invariant Neural-ODE formulation to constrain the solution space and can be trained on datasets with missing data and variable time intervals between the visits. We demonstrated two approaches to constrain the solution to physiologically plausible ones and introduced temporal Dice loss for training with longitudinal data. The state of the art performance of our method on the two distinct tasks of predicting the progression of GA in retinal OCT and the ventricles in brain MRI demonstrates its efficacy and generalizability across different anatomical structures and imaging modalities.

We focused on the task of providing predictions from a single scan, targeting the scenario of patient recruitment for clinical trials, where typically only a baseline scan is available. In addition to enriching for fast progressors, our method could serve as a control digital twin of a patient receiving a treatment to predict the expected outcomes without the treatment, thereby reducing the need for costly sham arms in clinical trials. Further extension of this work to utilize multiple prior scans and update the predictions as the new data becomes available offers a promising direction for future work.

\bibliographystyle{IEEE}
\bibliography{bibtex}

\end{document}


\maketitle

\begin{table*}[t]
\centering
\caption{Comparison of the related work to the proposed method. }
\label{tab_comp}
\centering
\resizebox{\textwidth}{!}{\begin{tabular}{|c|c|c|c|c|c|c|c|}
\hline
\thead{Method} & \thead{Base methods} & \thead{Data} & \thead{Application} & \thead{Irregular\\time series} & \thead{Interpolation/\\Extrapolation} & \thead{Segmentation\\metrics} & \thead{Progression\\metrics} \\
\hline
\makecell{Wang et al. \cite{Wang2018}} &  LSTM \cite{lstm} & \makecell{Tabular data}  & \makecell{AD progression} & \makecell{+} & +/+ & - & Accuracy  \\
\hline
\makecell{Nguyen et al. \cite{NGUYEN2020117203}} &  MinimalRNN\cite{MinimalRNNTM} & \makecell{Tabular data}  & \makecell{AD diagnosis,\\ biomarkers} & \makecell{$\pm$\\(interp.)} &  -/+ & - & \makecell{AUROC,\\MAE}  \\
\hline
\makecell{Ismail et al \cite{tadpole_melba}} &  LSTM\cite{lstm} & \makecell{Tabular data}  & \makecell{AD progression} & \makecell{+} & +/+ & - & MAE  \\
\hline
\makecell{Hill et al \cite{tadpole_melba}} &  Regression & \makecell{Tabular data}  & \makecell{AD progression} & \makecell{+} & +/+ & - & MAE  \\
\hline
\makecell{Venkatraghavan et al \cite{tadpole_melba}} &  \makecell{Linear Mixed \\Effects Model} & \makecell{Tabular data}  & \makecell{AD progression} & \makecell{+} & +/+ & - & MAE  \\
\hline
\makecell{Ezhov et al. \cite{Ezhov2019NeuralPE}} &  \makecell{ MDN \cite{Bishop94mixturedensity} \\ growth model} & \makecell{MRI}  & \makecell{Brain tumor\\ growth} & + & -/- & - & -  \\
\hline
\makecell{Petersen et al. \cite{Petersen2019DeepPM}} &  \makecell{ UNet \cite{unet}, \\ ELBO} & \makecell{MRI}  & \makecell{Brain tumor\\ growth} & - & -/- & \makecell{Query Volume\\Dice \cite{Petersen2019DeepPM}} & -  \\
\hline
\makecell{Petersen et al. \cite{dgg}} &  \makecell{ UNet \cite{unet}, \\ Attention \cite{transformers}} & \makecell{MRI}  & \makecell{Brain tumor\\ growth} & + & +/+ & \makecell{Query Volume\\Dice \cite{Petersen2019DeepPM}} & -  \\
\hline
\makecell{Hao et al. \cite{Hao}} & NeuralODE \cite{node} & \makecell{MRI, PET}  & \makecell{Amyloid\\accumulation} & + & +/+ & - & \makecell{Comparison with\\linear trend} \\
\hline
\makecell{De Brouwer et al. \cite{Brouwer2020LongitudinalMO}} & GRU-ODE-Bayes \cite{Brouwer2019GRUODEBayesCM} & \makecell{Tabular data}  & \makecell{MS progression} & + & +/+ & - & \makecell{AUROC} \\

\hline
\makecell{Niu et al. \cite{Niu2016}} & \makecell{Random\\Forest} & \makecell{3D OCT}  & \makecell{GA progression} & -  & -/- & \makecell{Dice,\\Sens.,Spec.,\\Pearson's r} & Pearson's r \\
\hline
\makecell{Zhang et al. \cite{Zhang2019}} & FCN \cite{fcn} & \makecell{2D OCT\\projections}  & \makecell{GA progression} & - & -/- & \makecell{Dice,\\Overlap ratio} & - \\
\hline
\makecell{Zhang et al. \cite{ZHANG2021}} & \makecell{Bi-LSTM,\\UNet3D\cite{unet3d}} & \makecell{3D OCT}  & \makecell{GA progression\\from two visits} & + & +/+ & \makecell{Dice} & Pearson's r \\
\hline
\makecell{Gigon et al. \cite{taylor}} & FCNN & \makecell{2D OCT\\projections}& \makecell{GA progression} & + & +/+ & \makecell{Dice} & RMSE \\
\hline
\makecell{\bf{Proposed}} &  NeuralODE \cite{node} & \makecell{3D OCT\\
MRI}  & \makecell{GA progression\\AD progression}  & + & +/+ & Dice & $R^2$, Pearson's r \\
\hline
\end{tabular}}
\end{table*}

\bibliographystyle{IEEE}
\bibliography{bibtex}